\documentclass[lettersize,journal]{IEEEtran}
\usepackage{amsmath,amsfonts}
\usepackage{algorithmic}
\usepackage{algorithm}
\usepackage{array}
\usepackage[caption=false,font=normalsize,labelfont=sf,textfont=sf]{subfig}
\usepackage{textcomp}
\usepackage{stfloats}
\usepackage{url}
\usepackage{verbatim}
\usepackage{graphicx}
\usepackage{cite}
\usepackage{booktabs}
\usepackage{caption}
\usepackage{multicol}
\usepackage{multirow}
\usepackage{makecell}
\usepackage[table]{xcolor}  

\definecolor{best}{RGB}{0,150,0}      
\definecolor{third}{RGB}{30,124,235}  
\definecolor{second}{RGB}{220,120,0}  

\begin{document}

\title{\textit{SatelliteCalculator}: A Multi-Task Vision Foundation Model for Quantitative Remote Sensing Inversion}


\author{Zhenyu Yu$^{1}$, Mohd. Yamani Idna Idris$^{2,*}$, Pei Wang$^{3,*}$%

        %
        %
        \thanks{${^1}$Faculty of Computer Science and Information Technology, Universiti Malaya, Kuala Lumpur 50603, Malaysia (E-mail: yuzhenyuyxl@foxmail.com).}%
        \thanks{${^2}$Faculty of Computer Science and Information Technology, Universiti Malaya, Kuala Lumpur 50603, Malaysia (Email: yamani@um.edu.my).}%
        \thanks{${^3}$Faculty of Information Engineering and Automation, Kunming University of Science and Technology, Kunming 210098, China (Email: peiwang@kust.com).}%
        \thanks{${^*}$Corresponding author: Mohd. Yamani Idna Idris and Pei Wang.}
}

\markboth{Journal of \LaTeX\ Class Files,~Vol.~14, No.~8, April~2025}%
{Shell \MakeLowercase{\textit{et al.}}: A Sample Article Using IEEEtran.cls for IEEE Journals}

\IEEEpubid{0000--0000/00\textdollar00.00~\copyright~2025 IEEE}

\maketitle

\begin{abstract}
Quantitative remote sensing inversion plays a critical role in environmental monitoring, enabling the estimation of key ecological variables such as vegetation indices, canopy structure, and carbon stock. Although vision foundation models have achieved remarkable progress in classification and segmentation tasks, their application to physically interpretable regression remains largely unexplored. Furthermore, the multi-spectral nature and geospatial heterogeneity of remote sensing data pose significant challenges for generalization and transferability. To address these issues, we introduce \textit{SatelliteCalculator}, the \textbf{first vision foundation model tailored for quantitative remote sensing inversion}. By leveraging physically defined index formulas, we automatically construct a large-scale dataset of over \textbf{one million paired samples} across eight core ecological indicators. The model integrates a frozen Swin Transformer backbone with a prompt-guided architecture, featuring cross-attentive adapters and lightweight task-specific MLP decoders. Experiments on the Open-Canopy benchmark demonstrate that \textit{SatelliteCalculator} achieves competitive accuracy across all tasks while significantly reducing inference cost. Our results validate the feasibility of applying foundation models to quantitative inversion, and provide a scalable framework for task-adaptive remote sensing estimation. The code and dataset are available at: https://github.com/YuZhenyuLindy/SatelliteCalculator
\end{abstract}

\begin{IEEEkeywords}
Quantitative Remote Sensing, Foundation Model, Multi-Task, Inversion, Estimation.
\end{IEEEkeywords}


\section{Introduction}
\label{sec:introduction}

Remote sensing inversion plays a vital role in monitoring terrestrial ecosystems by enabling the quantitative estimation of key environmental variables such as vegetation indices, aboveground biomass, and canopy structure~\cite{running1999remote, asner2011carbon,yu2023impact}. Traditional approaches rely on physical models and empirical formulas, which require expert knowledge, handcrafted features, and often suffer from limited regional generalization~\cite{liang2004quantitative,tan2023spatiotemporal}. With the rapid advancement of computer vision and deep learning, visual models have shown promise in various remote sensing tasks~\cite{zhu2017deep,yu2024capan}. However, most existing efforts are restricted to classification, segmentation, or change detection, leaving the domain of \textit{quantitative inversion} largely underexplored~\cite{li2022deep,wang2024multi}.

Unlike natural images, remote sensing data contains multiple spectral bands beyond standard RGB, presenting unique challenges to model design~\cite{ayush2021geography,yu2025yuan}. These include adapting input structures, transferring pretrained weights, and maintaining consistency across tasks. In this work, we use four high-resolution Sentinel-2 bands (blue, green, red, and near-infrared), which capture key spectral information for ecological estimation~\cite{wang2023remotebenchmark, ru2023learning,yu2025qrs}. Effectively adapting vision models to multi-band inputs while ensuring stability and accuracy remains a core technical bottleneck.

To address this, we explore the feasibility of applying deep models to quantitative remote sensing inversion. Leveraging physically defined formulas of spectral and structural indices, we construct a large-scale supervised dataset—containing over one million paired samples—spanning eight core ecological variables~\cite{tucker1979red, liang2004quantitative,yu2025ai}. This approach alleviates the annotation bottleneck~\cite{zhu2017deep,yu2024improved}, and enables the training of task-specific, interpretable models beyond conventional expert-driven inversion~\cite{song2021deep, li2022deep}.

While foundation models have emerged in remote sensing for tasks such as classification, atmospheric retrieval, or general-purpose understanding~\cite{xu2023rsfm}, none are tailored to \textit{quantitative inversion}. This setting requires regression accuracy across diverse physical targets and geographic regions~\cite{li2022deep}.

We propose \textit{SatelliteCalculator}, a multi-task vision foundation model for quantitative remote sensing inversion. The model jointly estimates eight ecological indicators, including NDVI, GNDVI, SAVI, EVI, NDWI, canopy height (H), aboveground biomass (AGB), and carbon stock (CS), from high-resolution Sentinel-2 imagery. Trained on physically consistent data derived from the Open-Canopy dataset, it demonstrates strong performance and generalization across tasks.

Our key \textbf{contributions} are as follows:
\begin{itemize}
    \item \textbf{Quantitative inversion foundation model.} We propose \textit{SatelliteCalculator}, the first vision foundation model specifically designed for quantitative remote sensing inversion, jointly estimating eight ecological variables under a unified multi-task framework.

    \item \textbf{Prompt-guided cross-attentive adapter.} We introduce a lightweight task adaptation module that injects prompt-conditioned semantics into shared Swin features via cross-attention, enabling flexible and accurate multi-task regression.

    \item \textbf{Task-Specific Decoder.} We adopt minimal MLP decoders for each task head, achieving a strong balance between accuracy and inference efficiency with low memory overhead.
\end{itemize}

To support scalable training, we construct a large-scale dataset by synthesizing over one million samples based on spectral and structural definitions, aligned with physical remote sensing principles.

\section{Related Work}
\label{sec:relatedWork}

\subsection{Traditional Approaches}

Quantitative remote sensing inversion has traditionally relied on physical modeling and empirical regression techniques to estimate ecological indicators such as NDVI, aboveground biomass (AGB), and canopy height (H)~\cite{tucker1979red, running1999remote, lemaire2008canopy}. Radiative transfer models (RTMs), such as PROSAIL and SAIL, simulate canopy reflectance using radiative transport equations, offering physically interpretable but computationally intensive solutions. In contrast, index-based approaches construct vegetation indices (e.g., NDVI, EVI, SAVI, NDWI) from spectral band combinations and apply regression techniques—such as PCR and PLSR—for variable estimation~\cite{asner2011carbon, liu2020review}.

Despite their interpretability, these methods heavily depend on expert knowledge and handcrafted features, and often suffer from poor generalization across regions and sensors. Moreover, they are limited to single-variable estimation and cannot scale to modern multi-task, multi-source inversion scenarios.

\subsection{Deep Learning Approaches}

Recent advances in deep learning have enabled end-to-end estimation of geophysical variables from satellite imagery. Early works applied CNNs and encoder-decoder architectures (e.g., UNet) for tasks such as NDVI prediction, biomass estimation, and canopy height regression~\cite{zhang2018cnn, fang2019aboveground, song2021deep}. For temporal modeling, LSTMs and RNNs have been used to capture vegetation dynamics in time-series data~\cite{zhao2019lstm, chen2020deepcrop}. More recently, Transformer-based architectures have gained popularity for their global attention capabilities and effectiveness in handling multi-band inputs~\cite{li2022deep, yuan2021s2t}.

However, most of these models are designed for single-task settings and customized for specific indicators or regions. Additionally, many architectures are inherited from natural image models and require manual adaptation to accommodate remote sensing inputs with multiple spectral bands. As a result, current models struggle with \textit{task scalability, spectral adaptability, and generalization}, which are essential for unified inversion systems.

\subsection{Foundation Models in Remote Sensing}

Vision foundation models such as MAE~\cite{he2022masked}, DINO~\cite{caron2021emerging}, and SAM~\cite{kirillov2023segment} have demonstrated strong generalization and transferability in natural image tasks. Inspired by these successes, recent studies have introduced remote sensing-specific foundation models, such as SatMAE~\cite{dong2023satmae}, RS-FM~\cite{xu2023rsfm}, and RemoteCLIP~\cite{wang2023remotebenchmark}, targeting classification, retrieval, or contrastive pretraining. GeoGPT~\cite{lu2023geogpt} further explores vision-language alignment for geospatial understanding.

While these models support discrete tasks and multi-modal learning, they lack support for \textit{quantitative inversion}, which requires accurate regression of continuous, physically grounded variables. Compared to classification, inversion involves different loss functions, output spaces, and supervision forms. It also requires precise handling of multi-spectral or hyper-spectral inputs and adaptation to heterogeneous ecological indicators.

Despite recent progress, there remains a critical gap in foundation models that jointly support: (1) \textbf{multi-band spectral adaptation}, (2) \textbf{physically interpretable regression}, and (3) \textbf{scalable multi-task learning}. To bridge this gap, we propose \textit{SatelliteCalculator}, the \textbf{first vision foundation model} specifically designed for quantitative remote sensing inversion.

\section{Method}
\label{sec:method}

\subsection{Overview}

We formulate quantitative remote sensing inversion as a prompt-guided multi-task regression problem. Given a four-band Sentinel-2 VHR image $X \in \mathbb{R}^{H \times W \times 4}$ and a task prompt $P_t$ that specifies the target variable (e.g., NDVI, AGB), the objective is to predict a spatially continuous response map $\hat{Y}_t \in \mathbb{R}^{H \times W}$ reflecting the corresponding ecological quantity.

\begin{figure*}
    \centering
    \includegraphics[width=1.0\linewidth]{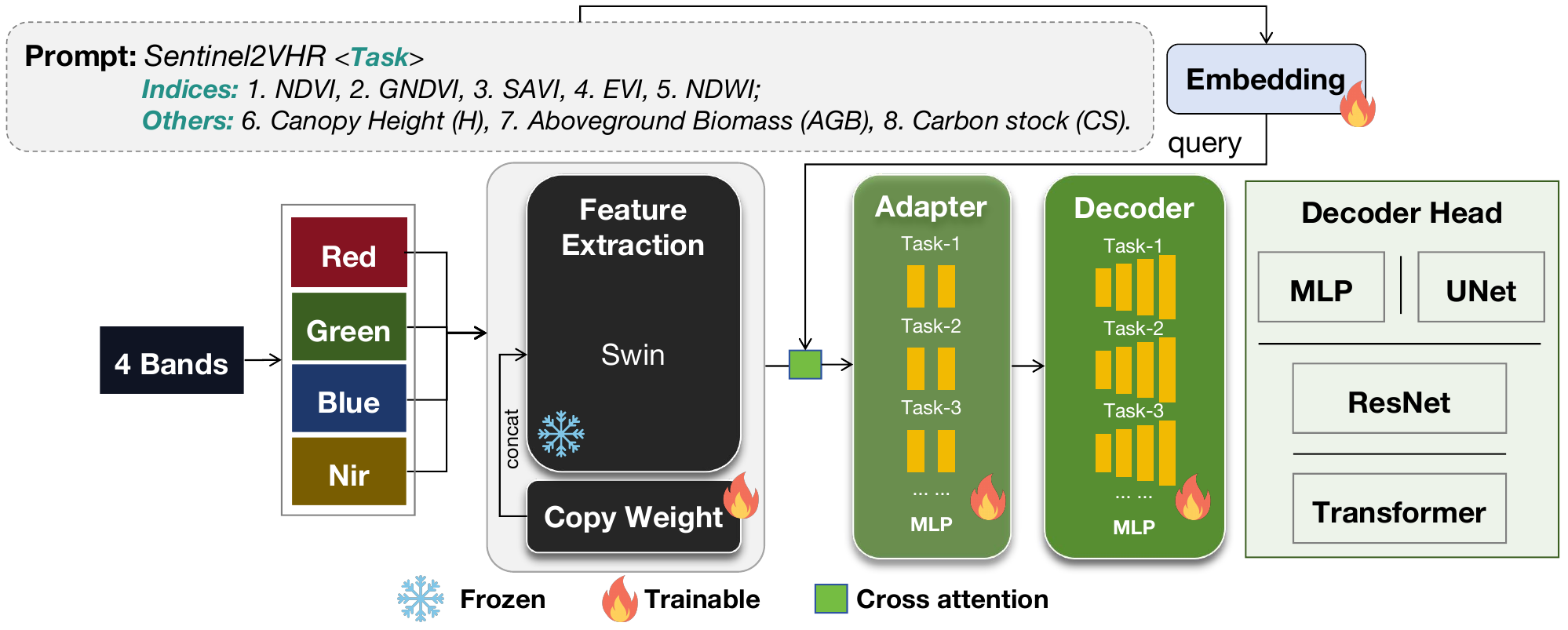}
    \caption{Overview of the \textit{SatelliteCalculator} framework. The model takes a multi-band Sentinel-2 VHR image and a task prompt as input, and performs quantitative inversion through four components: (1) prompt embedding module, (2) Swin Transformer feature extractor, (3) Cross-attentive adapter, and (4) Task-specific decoder.}
    \label{fig_overview}
\end{figure*}

The proposed framework, \textit{SatelliteCalculator}, consists of four modular components (Fig.~\ref{fig_overview}):  
\begin{enumerate}
    \item a \textbf{prompt embedding module} that converts each task prompt $P_t$ into a learnable query vector to encode task semantics;
    \item a \textbf{Swin Transformer feature extractor} that encodes shared visual representations from the input image $X$;
    \item a \textbf{cross-attentive adapter} that injects task-specific information into the shared features via cross-attention between the prompt query and image tokens;
    \item a \textbf{task-specific decoder} implemented as a multi-layer perceptron (MLP), designed to provide flexible regression capacity for each task without modifying the shared feature extractor.
\end{enumerate}

This architecture decouples shared visual representation learning from task-specific adaptation, enabling scalable, efficient, and generalizable multi-task remote sensing inversion.

\subsection{Prompt Embedding Module}

To enable flexible and scalable multi-task inversion across diverse ecological variables, we introduce a task prompt mechanism that explicitly encodes the semantics of each estimation target. Formally, each task \( t \in \mathcal{T} \) is identified by a discrete token \( P_t \) (e.g., \texttt{"NDVI"}, \texttt{"AGB"}), representing the variable to be predicted.

We implement a learnable prompt embedding layer that maps each token \( P_t \) to a continuous query vector \( q_t \in \mathbb{R}^d \), referred to as the task query:
\begin{equation}
q_t = \text{PromptEmbed}(P_t)
\end{equation}
Here, \( \text{PromptEmbed} \) is a shared embedding matrix of size \( |\mathcal{T}| \times d \), where \( |\mathcal{T}| \) denotes the number of supported tasks and \( d \) is the embedding dimension aligned with the visual token features from the backbone.

The resulting query vector \( q_t \) acts as a high-level controller, conditioning downstream modules (e.g., cross-attentive adapters and task-specific decoders) on the nature of the target variable. This design ensures that the shared visual features can be dynamically adapted to different tasks through soft attention mechanisms guided by task semantics.

Notably, the prompt embedding parameters are fully trainable~\includegraphics[height=1em]{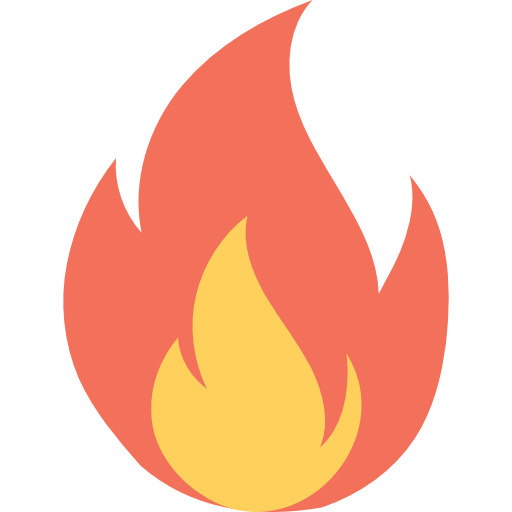} and are optimized jointly with the rest of the model under a unified multi-task loss. This formulation provides a flexible interface for expanding the model to new inversion targets, simply by appending additional prompt entries to the embedding table without retraining the entire network.

In summary, the prompt embedding module introduces semantic task-awareness into the model, enabling modular, interpretable, and parameter-efficient multi-task learning for quantitative remote sensing inversion.

\subsection{Swin Transformer Feature Extractor}

To obtain high-resolution spatial representations from multi-spectral inputs, we adopt a Swin Transformer as the visual backbone for feature extraction. Given an input image \( X \in \mathbb{R}^{H \times W \times 4} \), where the last dimension corresponds to the four selected spectral bands from Sentinel-2 (blue, green, red, and near-infrared), the backbone processes the image into a sequence of patch embeddings:
\begin{equation}
F = \text{Swin}(X), \quad F \in \mathbb{R}^{N \times d}, \quad N = \frac{H \cdot W}{p^2}
\end{equation}
Here, \( p \) denotes the patch size, \( N \) is the total number of non-overlapping patches, and \( d \) is the feature embedding dimension. The output \( F \) is a set of task-agnostic visual tokens encoding local and global spatial patterns.

The Swin Transformer introduces a hierarchical design based on windowed self-attention, which significantly reduces computation while preserving contextual information. Specifically, it employs a shifted window scheme to capture both short- and long-range dependencies across patch neighborhoods, making it well-suited for capturing structured spatial features in high-resolution remote sensing imagery.

In this framework, the Swin Transformer is initialized from weights pretrained on ImageNet-21k and is kept entirely frozen during training \includegraphics[height=1em]{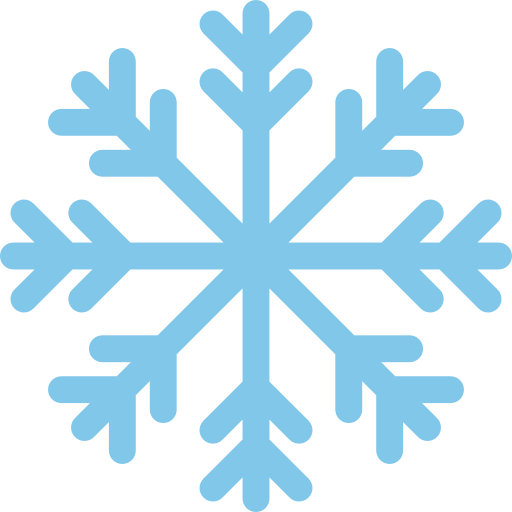}. Freezing the backbone serves two purposes: (1) it leverages powerful general-purpose visual priors, which are transferable to geospatial domains despite spectral domain shifts, and (2) it prevents overfitting, especially given the limited diversity of geophysical supervision signals in downstream tasks.

The extracted feature tensor \( F \) is shared across all target tasks and is further modulated through task-specific adaptation layers. By decoupling general visual encoding from downstream regression heads, this design ensures that the model maintains a unified feature space while supporting task-level specialization through modular conditioning.

\subsection{Cross-Attentive Adapter}

To enable flexible task-specific modulation of shared visual features, we introduce a dedicated \textit{cross-attentive adapter} for each target task \( t \in \mathcal{T} \). The adapter is responsible for infusing task semantics into the visual representation by conditioning the shared features \( F \in \mathbb{R}^{N \times d} \) on the corresponding task prompt vector \( q_t \in \mathbb{R}^d \).

Formally, the adapter module consists of two main stages: cross-attention and a task-specific feedforward transformation. First, the query vector \( q_t \) is broadcasted and used as the query in a cross-attention layer, while the visual tokens \( F \) serve as the key and value matrices:
\begin{equation}
F^{\text{attn}}_t = \text{CrossAttn}(q_t, F), \quad F^{\text{attn}}_t \in \mathbb{R}^{N \times d}
\end{equation}
This operation enables the model to dynamically select relevant spatial features based on the semantic intent of the task. The output of the attention mechanism is then passed through a task-specific multi-layer perceptron (MLP) to produce the adapted feature map:
\begin{equation}
F_t = \text{Adapter}_t(F, q_t) = \text{MLP}_t\left(F^{\text{attn}}_t\right), \quad F_t \in \mathbb{R}^{N \times d}
\end{equation}

Each adapter \( \text{Adapter}_t \) is independently parameterized and optimized \includegraphics[height=1em]{figs/flame.png}, allowing the model to disentangle task-specific behavior while sharing a frozen backbone. The modular nature of this design facilitates extensibility to new tasks by simply adding new adapters, without altering the core architecture.

Moreover, this cross-attentive design provides a mechanism for \textit{task-conditioned feature extraction}, where each ecological variable (e.g., NDVI, AGB, CS) learns to focus on its own relevant spectral and spatial patterns while leveraging a common visual foundation. Such a formulation ensures both model generality and task specialization—two critical properties for scalable remote sensing inversion.

\subsection{Task-Specific Decoder}

For each task \( t \in \mathcal{T} \), we introduce a dedicated decoder \( \text{Decoder}_t \) that transforms the task-adapted features \( F_t \in \mathbb{R}^{N \times d} \) into a dense spatial prediction map \( \hat{Y}_t \in \mathbb{R}^{H \times W} \), representing the estimated geophysical quantity (e.g., NDVI, AGB, CS). Each decoder is implemented as a task-specific multi-layer perceptron (MLP) \includegraphics[height=1em]{figs/flame.png}, independently parameterized and optimized to capture the statistical characteristics, value ranges, and units of the corresponding target variable.

The decoder operates on patch-level tokens and maps them back to the spatial domain through a sequence of linear projections and non-linear activations. In practice, this involves reshaping the output from \( N = H \cdot W / p^2 \) patch tokens to a \( H \times W \) grid via inverse patch embedding or interpolation, depending on the resolution alignment strategy used.

Formally, the decoding process is defined as:
\begin{equation}
\hat{Y}_t = \text{Decoder}_t(F_t), \quad \hat{Y}_t \in \mathbb{R}^{H \times W}
\end{equation}

To train the model across all tasks in a unified manner, we adopt a weighted mean absolute error (MAE) objective:
\begin{equation}
\mathcal{L} = \sum_{t=1}^{T} \lambda_t \cdot \text{MAE}(\hat{Y}_t, Y_t)
\end{equation}
where \( Y_t \) denotes the ground truth target map for task \( t \), and \( \lambda_t \) is a task-specific scalar weight controlling its contribution to the total loss. The weights can be heuristically assigned or estimated based on prior statistics such as value range or initial loss magnitude to ensure balanced optimization across heterogeneous tasks.

During training, only the prompt embedding layer, cross-attentive adapters, and task-specific decoders are updated \includegraphics[height=1em]{figs/flame.png}. The Swin Transformer feature extractor remains frozen \includegraphics[height=1em]{figs/snowflake.png} to preserve general visual priors and enhance training stability. This selective optimization strategy reduces the number of trainable parameters, facilitates transferability, and ensures efficient adaptation to new tasks or domains.

\begin{algorithm}[t]
\caption{\textit{SatelliteCalculator} Inversion Framework}
\label{alg:satellitecalculator}
\begin{algorithmic}[1]
\REQUIRE Multi-band image $X \in \mathbb{R}^{H \times W \times 4}$, task set $\mathcal{T}$ with prompts $\{P_t\}_{t=1}^T$, task weights $\{\lambda_t\}_{t=1}^T$
\ENSURE Output predictions $\{\hat{Y}_t\}_{t=1}^T$ for all tasks $t \in \mathcal{T}$

\vspace{0.5em}
\textcolor{gray}{\# (1) Prompt Embedding Module}
\FOR{each task $t \in \mathcal{T}$}
    \STATE $q_t \gets \textsc{PromptEmbed}(P_t)$ \hfill \textcolor{gray}{// Encode task semantics as query vector $q_t \in \mathbb{R}^d$}
\ENDFOR

\vspace{0.5em}
\textcolor{gray}{\# (2) Swin Transformer Feature Extractor}
\STATE $F \gets \textsc{SwinTransformer}(X)$ \hfill \textcolor{gray}{// Extract task-agnostic visual features $F \in \mathbb{R}^{N \times d}$}

\vspace{0.5em}
\textcolor{gray}{\# (3) Cross-Attentive Adapter and (4) Task-Specific Decoder}
\FOR{each task $t \in \mathcal{T}$}
    \STATE $F_t \gets \textsc{CrossAttn}(F, q_t)$ \hfill \textcolor{gray}{// Fuse prompt $q_t$ with visual features}
    \STATE $F_t \gets \textsc{MLP}_t(F_t)$ \hfill \textcolor{gray}{// Apply task-specific transformation}
    \STATE $\hat{Y}_t \gets \textsc{Decoder}_t(F_t)$ \hfill \textcolor{gray}{// Output prediction map $\hat{Y}_t \in \mathbb{R}^{H \times W}$}
\ENDFOR

\vspace{0.5em}
\textcolor{gray}{\# (5) Multi-task Optimization}
\STATE $\mathcal{L} \gets 0$
\FOR{each task $t \in \mathcal{T}$}
    \STATE $\mathcal{L}_t \gets \textsc{MAE}(\hat{Y}_t, Y_t)$ \hfill \textcolor{gray}{// Compute task loss}
    \STATE $\mathcal{L} \gets \mathcal{L} + \lambda_t \cdot \mathcal{L}_t$
\ENDFOR

\vspace{0.5em}
\RETURN $\{\hat{Y}_t\}_{t=1}^T$, total loss $\mathcal{L}$
\end{algorithmic}
\end{algorithm}

\section{Experiment}
\label{sec:experiment}

\subsection{Data Description}



We utilize the Open-Canopy dataset \cite{fogel2024open}, which provides high-quality remote sensing data for forest monitoring. Open-Canopy covers over 87,000~$km^2$ across France and offers 1.5~$m$ very high-resolution (VHR) satellite imagery alongside airborne LiDAR-derived canopy height maps, enabling detailed structural analysis of open forest areas.
To enable multi-task quantitative inversion, we construct a large-scale dataset based on Open-Canopy imagery. Specifically, we use spectral reflectance from four Sentinel-2 bands (B2–B4, B8) to compute five vegetation and water indices—NDVI, GNDVI, EVI, SAVI, and NDWI—using standard physical formulas. Structural and biomass-related labels are derived by directly using the provided canopy height (H) from LiDAR data, while aboveground biomass (AGB) and carbon stock (CS) are estimated through regionally calibrated empirical models.
The final dataset contains approximately \textbf{one million paired samples}, each consisting of a four-band input image and its corresponding multi-task ground truth targets. This large-scale, physically consistent dataset enables supervised training for accurate and generalizable multi-task remote sensing inversion. More details are provided in Appendix~\ref{sec_appendix_dataset}.

\subsection{Experimental Settings}
The experiments were conducted on 8 NVIDIA A100 GPU with 80 GB of memory. The model training employs the \(\ell_1\) norm as the loss function to enhance the stability of height prediction. In data processing, training samples are randomly cropped to \( 224 \times 224 \) pixels and undergo data augmentation, including scaling (0.5 to 2 times) and rotation (0°, 90°, 180°, 270°), to improve the model’s generalization ability. During optimization, the Adam optimizer is used with a learning rate of \( 10^{-4} \), combined with the ReduceLROnPlateau scheduler (patience of 1, decay factor of 0.5) to dynamically adjust the learning rate. Additionally, an early stopping strategy with a patience of 3 is applied to prevent overfitting.

\subsection{Evaluation Metrics}
We evaluate canopy height prediction performance using seven metrics: \textbf{Root Mean Square Error (RMSE)} \cite{hodson2022root}, \textbf{Mean Absolute Error (MAE)} \cite{hodson2022root}, \textbf{Normalized MAE (nMAE)} \cite{park2006naive}, \textbf{Bias} \cite{sterne2005regression}, \textbf{Tree Cover IoU} \cite{timilsina2020mapping}, \textbf{Coefficient of Determination (\(\mathbf{R^2}\))} \cite{nagelkerke1991note}, and \textbf{Peak Signal-to-Noise Ratio (PSNR)} \cite{hore2010image}. RMSE and MAE respectively capture the squared and absolute differences between predicted and ground truth values, with RMSE being more sensitive to large errors. nMAE normalizes the absolute error by the target height and is computed only on pixels with ground truth values above 2 $m$. Bias measures the signed average error, capturing systematic over- or under-estimation. The \(R^2\) score evaluates the proportion of variance explained by the model, with higher values indicating stronger predictive performance. PSNR measures the logarithmic ratio between the signal and reconstruction noise, reflecting global fidelity. Tree Cover IoU is calculated by thresholding predicted and ground truth height maps at 2 $m$ to produce binary vegetation masks and computing the intersection-over-union between them. All metrics are computed within the vegetation mask and restricted to pixels with ground truth height below 60 $m$.

\subsection{Comparison}

\subsubsection{Single-Task Backbone Comparison}
To evaluate the performance of different backbone architectures in quantitative remote sensing tasks, we conduct a systematic comparison of eight single-task models on the Open-Canopy benchmark dataset. Each model is independently trained to regress pixel-wise canopy height from Sentinel-2 VHR imagery. The compared models include convolutional networks (UNet, DeepLabv3), several mainstream vision transformers (ViT-B, HVIT, PCPVT, SWIN, PVTv2), and a self-supervised foundation model (ScaleMAE).

The main goal of this experiment is to identify the most suitable backbone for quantitative remote sensing inversion. As shown in Table~\ref{table_single_task}, transformer-based models consistently outperform convolutional ones across all evaluation metrics. For instance, the RMSE of UNet and DeepLabv3 reaches 4.18 $m$ and 4.83 $m$, respectively, while SWIN and PVTv2 reduce the RMSE to 4.00 $m$ and 4.02 $m$, demonstrating better stability and lower error. PVTv2 also achieves the best MAE (2.52 $m$) and zero bias, while SWIN ranks first or second in terms of nMAE (22.8\%) and Tree Cover IoU (90.5\%).

\begin{figure*}
    \centering
    \includegraphics[width=1\linewidth]{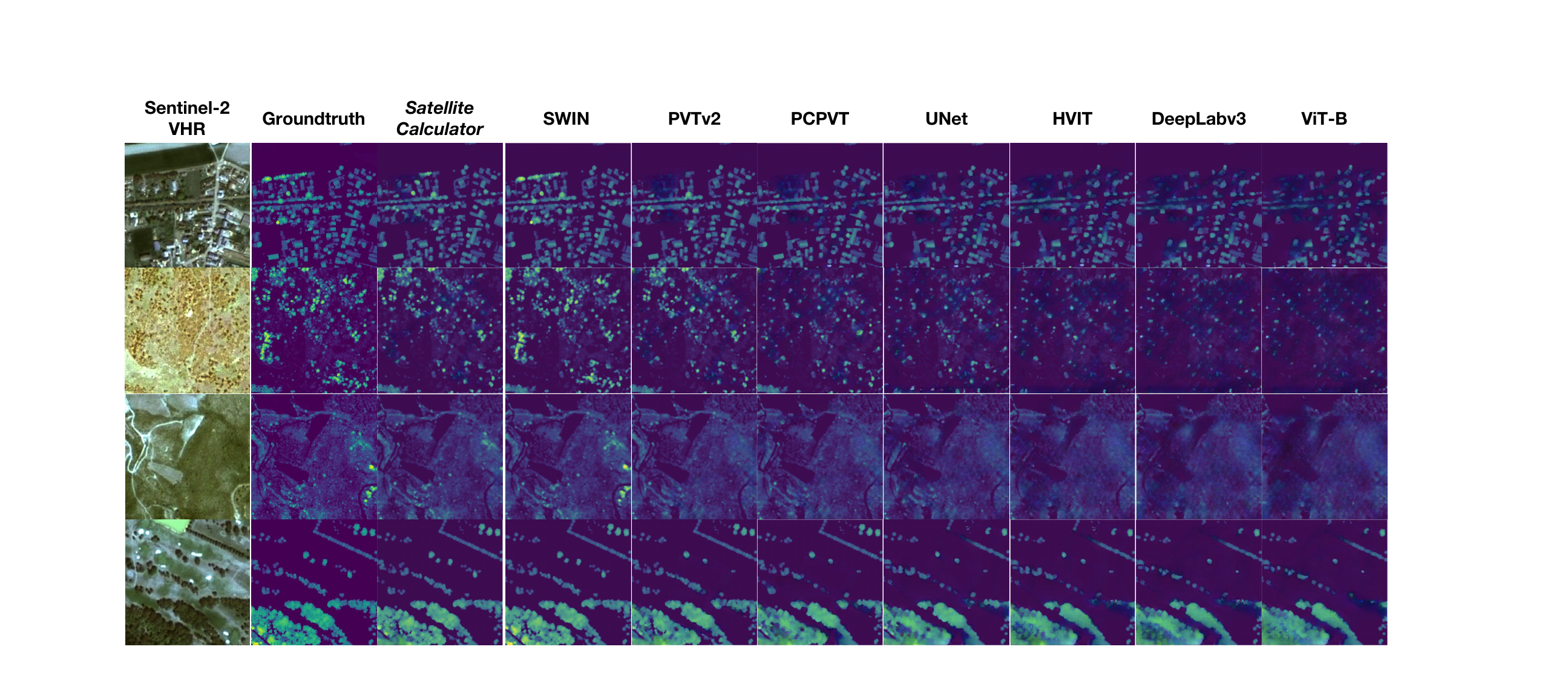}
    \caption{Comparison of canopy height inversion results across different models. From left to right: Sentinel-2 VHR input, groundtruth, estimations from \textit{SatelliteCalculator}, SWIN, PVTv2, PCPVT, UNet, HVIT, DeepLabv3, and ViT-B.}
    \label{fig_single_task}
\end{figure*}

Considering accuracy, robustness, and generalization, we adopt SWIN as the feature extractor backbone for our \textit{SatelliteCalculator}. The qualitative results in Figure~\ref{fig_single_task} further support this decision: compared with other models, predictions from SWIN and \textit{SatelliteCalculator} exhibit better edge preservation and spatial continuity, closely aligning with actual forest structure.

Notably, although \textit{SatelliteCalculator} is a multi-task vision foundation model designed to estimate eight types of ecological variables, its performance on canopy height prediction is comparable to the best single-task models such as SWIN and PVTv2 (e.g., MAE of 2.55 $m$, RMSE of 4.02 $m$), and it reaches a Tree Cover IoU of 90.5\%, only slightly below the Satlas-pretrained model (90.6\%). These results demonstrate that our proposed foundation model not only supports cross-task representation learning, but also maintains high accuracy on individual tasks, validating the feasibility and effectiveness of deep models in quantitative remote sensing inversion.

\begin{table*}[!ht]
    \centering
    \caption{Canopy height inversion performance of different backbone models. Color indicates: \textcolor{best}{\textbf{best}}, \textcolor{second}{\textbf{second-best}}, \textcolor{third}{\textbf{third-best}} across each metric.}
    \begin{tabular}{l l c c c c c}
    \toprule
    \textbf{Model} & \textbf{Pretraining} & \textbf{MAE (m)}$\downarrow$ & \textbf{nMAE (\%)}$\downarrow$ & \textbf{RMSE (m)}$\downarrow$ & \textbf{Bias (m)} & \textbf{Tree cov. IoU (\%)}$\uparrow$ \\
    \midrule
    UNet          & ImageNet1k & 2.67 & 23.8 & 4.18 & -0.30 & \textcolor{third}{\textbf{90.4}} \\
    DeepLabv3     & ImageNet1k & 3.18 & 28.4 & 4.83 & -0.26 & 88.0 \\
    ViT-B         & ImageNet21k & 4.26 & 37.8 & 6.06 & -0.84 & 86.0 \\
    HVIT          & ImageNet21k & 2.65 & 24.0 & 4.18 & -0.13 & 90.2 \\
    PCPVT         & ImageNet1k & \textcolor{third}{\textbf{2.57}} & \textcolor{third}{\textbf{23.1}} & \textcolor{third}{\textbf{4.06}} & -0.17 & \textcolor{third}{\textbf{90.4}} \\
    SWIN          & ImageNet21k & \textcolor{second}{\textbf{2.54}} & \textcolor{best}{\textbf{22.8}} & \textcolor{best}{\textbf{4.00}} & -0.11 & \textcolor{second}{\textbf{90.5}} \\
    PVTv2         & ImageNet1k & \textcolor{best}{\textbf{2.52}} & \textcolor{second}{\textbf{22.9}} & \textcolor{second}{\textbf{4.02}} & \textcolor{best}{\textbf{0.00}} & \textcolor{second}{\textbf{90.5}} \\
    \midrule
    ScaleMAE      & FotM & 3.45 & 31.2 & 5.13 & -0.48 & 88.2 \\
    ViT-B         & DINOV2 & 4.84 & 43.2 & 6.68 & -0.48 & 84.8 \\
    ViT-B         & CLIP\_OPENAI & 2.87 & 25.9 & 4.43 & \textcolor{third}{\textbf{-0.07}} & 89.7 \\
    ViT-L         & Tolan & 4.46 & 38.9 & 6.27 & -1.03 & 85.6 \\
    SWIN          & Satlas-pretrained & 2.56 & \textcolor{third}{\textbf{23.1}} & 4.09 & \textcolor{second}{\textbf{0.02}} & \textcolor{best}{\textbf{90.6}} \\
    \midrule
    \textbf{\textit{SatelliteCalculator}} & ImageNet21k & \textcolor{third}{\textbf{2.55}} & \textcolor{second}{\textbf{22.9}} & \textcolor{second}{\textbf{4.02}} & -0.13 & \textcolor{second}{\textbf{90.5}} \\
    \bottomrule
    \end{tabular}
    \label{table_single_task}
\end{table*}

\subsubsection{Loss Weighting Strategy}
In multi-task learning, the loss values of different tasks often vary in scale. Without proper regulation, tasks with higher loss magnitudes may dominate the training process, thereby hindering the optimization of other tasks. To address this issue, we adopt a fixed weighting strategy, where task-specific loss weights are pre-determined based on the relative magnitudes of initial training losses. This normalization ensures balanced optimization across all tasks during multi-task training.

Compared to dynamic weighting strategies—which may introduce instability or fluctuations during early training—fixed weights offer more stable optimization behavior. This is particularly suitable for our regression-based setting, where the target variables have well-defined physical meanings and heterogeneous ranges. For example, structural variables such as canopy height (H), aboveground biomass (AGB), and carbon stock (CS) typically have much larger value ranges than spectral indices (e.g., NDVI, EVI). Without appropriate rescaling, such structural tasks would dominate the total loss and skew the model’s convergence direction.

Specifically, we normalize the average loss magnitude of each task observed during the initial training phase to derive a fixed set of weights, as shown in Table~\ref{table_loss_weight}. Structural tasks (H, AGB, CS) receive higher weights (approximately 0.20–0.24), while spectral index tasks (NDVI, GNDVI, NDWI, etc.) are assigned lower weights (mostly around 0.04), reflecting the natural variation in their loss contributions. Experimental results show that this fixed weighting strategy provides stable performance and effectively prevents any single task from dominating the training process, thereby promoting balanced optimization across all tasks within a shared model architecture.


\begin{table}[!ht]
    \centering
    \caption{Loss weights assigned to each task during multi-task training. We adopt a fixed weighting strategy based on task-specific loss magnitudes to balance optimization across eight tasks.}
    \begin{tabular}{ccccc}
    \toprule
        \textbf{Task} & NDVI & GNDVI & SAVI & EVI \\ 
        \textbf{Loss weight} & 0.0386  & 0.0440  & 0.0501  & 0.1700  \\ \midrule
        \textbf{Task} & NDWI & H & AGB & CS \\ 
        \textbf{Loss weight} & 0.0418  & 0.2052  & 0.2121  & 0.2381  \\ 
        \bottomrule
    \end{tabular}
    \label{table_loss_weight}
\end{table}

\subsubsection{Task-wise Performance Analysis}

To comprehensively evaluate the difficulty of different ecological estimation tasks and the adaptability of our model, we conduct a task-wise analysis under the multi-task learning framework. The eight target variables include five spectral indices (NDVI, GNDVI, SAVI, EVI, NDWI) and three structural variables (canopy height (H), aboveground biomass (AGB), and carbon stock (CS)). Table~\ref{table_each_task_result} presents the single-task performance of each variable in terms of MAE, PSNR, \(R^2\), and RMSE.

Overall, spectral index tasks yield significantly better results than structural variables. For instance, the RMSEs of NDVI and GNDVI are as low as 0.22, with corresponding \(R^2\) scores exceeding 0.85, indicating highly accurate and stable predictions. In contrast, the RMSEs for AGB and CS reach 26.04\,t/ha and 21.84\,Mg/ha, respectively, with lower \(R^2\) values of 0.55 and 0.52. This performance gap is largely due to two factors: (1) spectral indices are physically well-defined and can be directly computed from the input bands; (2) structural variables depend not only on spectral reflectance but also on complex environmental and regional factors, making them inherently more difficult to model.

\begin{figure*}
    \centering
    \includegraphics[width=1\linewidth]{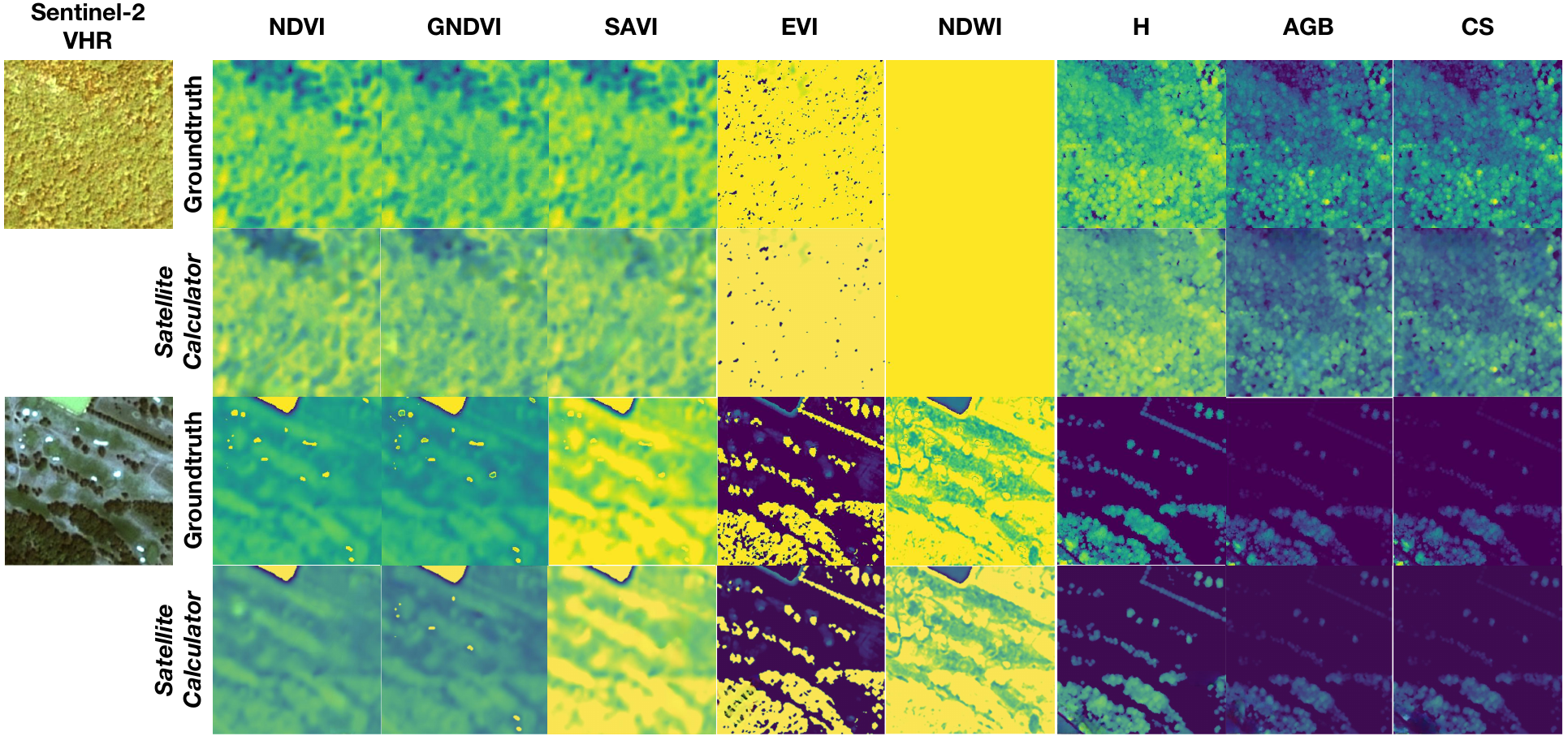}
    \includegraphics[width=1\linewidth]{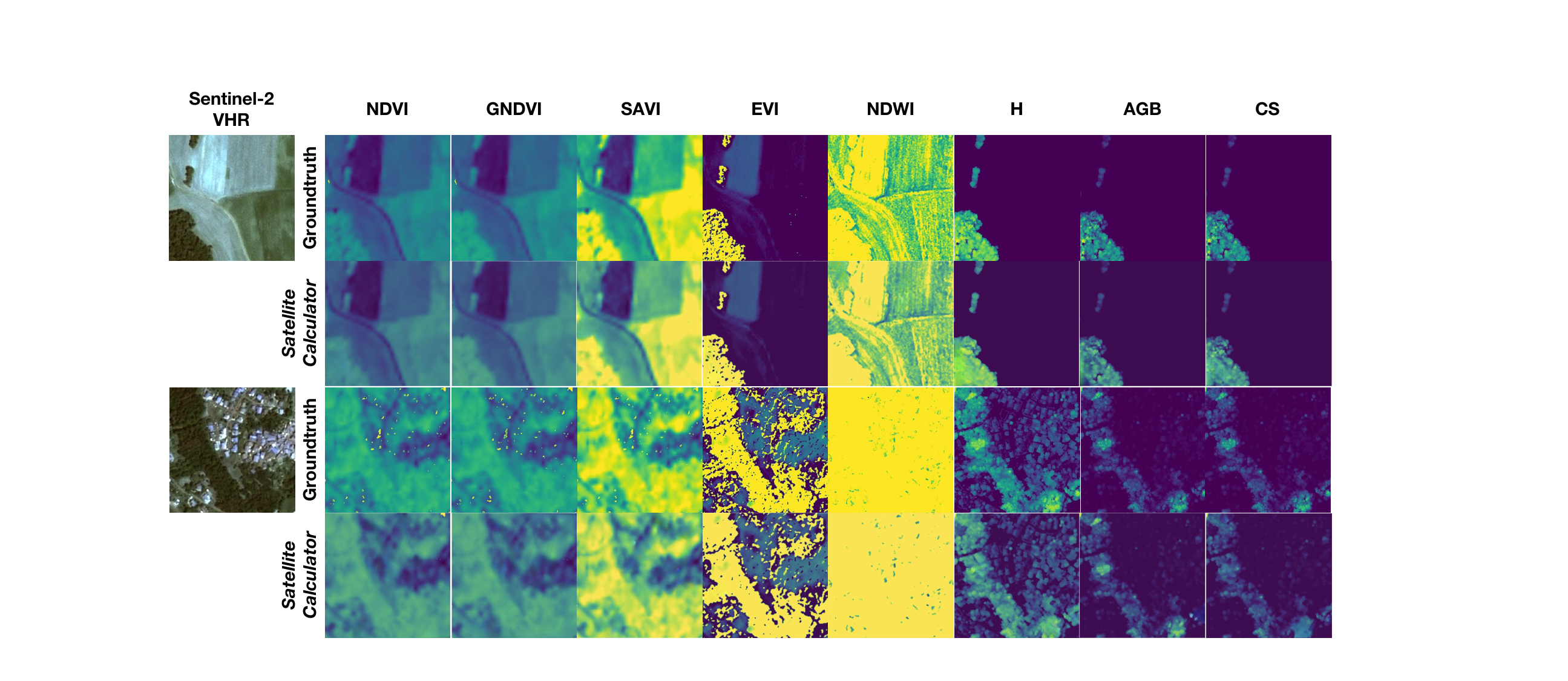}
    \caption{Multi-task inversion results on four Sentinel-2 scenes. We show groundtruth (top) and \textit{SatelliteCalculator} estimations (bottom) for five spectral indices and three structural variables.}
    \label{fig_each_task_result}
\end{figure*}

Figure~\ref{fig_each_task_result} shows qualitative results on four Sentinel-2 test scenes. The spectral index predictions from \textit{SatelliteCalculator} closely match the spatial distribution and boundary shapes of the ground truth, while the structural variables—especially AGB and CS—exhibit larger discrepancies in heterogeneous regions, such as forest edges or bare land areas.


\begin{table}[ht]
    \centering
    \caption{Results of \textit{SatelliteCalculator} on eight estimation tasks. We report MAE, PSNR, \(R^2\), and RMSE for five spectral indices (NDVI, GNDVI, SAVI, EVI, NDWI) and three structural variables (canopy height (H), aboveground biomass (AGB), and carbon stock (CS)).}
    \begin{tabular}{clcccc}
    \toprule
    \textbf{ID} & \textbf{Task} & \textbf{MAE}$\downarrow$ & \textbf{PSNR}$\uparrow$ & \(\mathbf{R^2}\)$\uparrow$ & \textbf{RMSE}$\downarrow$ \\
    \midrule
    1 & NDVI  & 0.05 & 29.27 & 0.85 & 0.22 \\
    2 & GNDVI & 0.06 & 28.99 & 0.86 & 0.22 \\
    3 & SAVI  & 0.07 & 27.24 & 0.77 & 0.27 \\
    4 & EVI   & 0.23 & 21.95 & 0.66 & 0.51 \\
    5 & NDWI  & 0.06 & 27.79 & 0.80 & 0.26 \\
    6 & H     & 2.55 (m)   & 21.05 & 0.61 & 4.02 (m) \\
    7 & AGB   & 21.34 (t/ha) & 21.37 & 0.55 & 26.04 (t/ha) \\
    8 & CS    & 16.80 (Mg/ha) & 21.12 & 0.52 & 21.84 (Mg/ha) \\
    \bottomrule
    \end{tabular}
    \label{table_each_task_result}
\end{table}

\subsection{Ablation Study}


\subsubsection{Effect of Decoder Architecture}

To evaluate the impact of different decoder architectures on multi-task remote sensing estimation, we compare four representative designs: {MLP, UNet, ResNet, and Transformer}. Decoder architecture is shown in Table \ref{tab_mlp_task_head}, \ref{tab_unet_task_head}, \ref{tab_resnet_task_head}, and \ref{tab_transformer_task_head}. {MLP consistently achieves more stable and accurate results in spectral index tasks} (Table~\ref{table_decoder_comparison}), attaining the best or second-best performance in terms of \(R^2\) and RMSE for NDVI, GNDVI, SAVI, and EVI. For structural variables such as AGB and CS, ResNet and UNet show slight advantages in specific metrics, but the overall differences from MLP are marginal. The Transformer decoder delivers competitive PSNR values, but suffers from unstable RMSE and \(R^2\) scores, indicating lower robustness when handling heterogeneous geospatial patterns.

\begin{figure}
    \centering
    \includegraphics[width=1\linewidth]{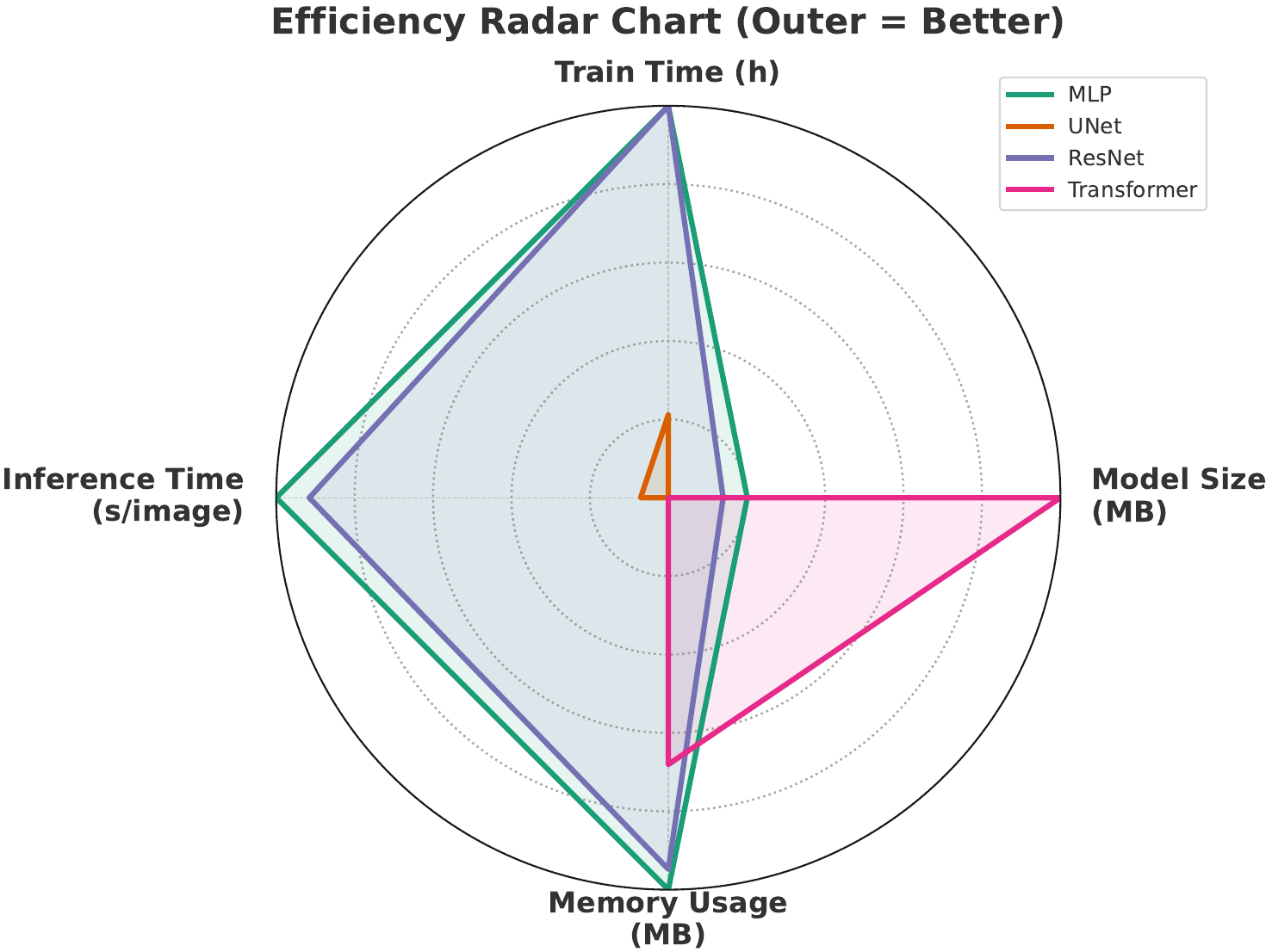}
    \caption{Efficiency comparison across four decoder architectures (MLP, UNet, ResNet, and Transformer). Each axis represents one of four evaluation metrics where outer regions indicate better performance. The chart highlights trade-offs between model complexity and efficiency, showing that MLP achieves optimal inference and memory usage.}
    \label{fig_efficiency}
\end{figure}

In terms of efficiency (Table~\ref{table_efficiency} and Figure~\ref{fig_efficiency}), {MLP significantly outperforms the other decoders}. The 4-layer MLP requires only {0.48 seconds per image} for inference and consumes just {1566 MB} of GPU memory—the lowest among all options. Its training time (19.51 hours) is nearly as fast as ResNet (19.49 hours), while both UNet and Transformer incur much higher costs in memory and runtime, which limits their practicality in large-scale or real-time applications.

Taking both accuracy and efficiency into account, {MLP achieves the best trade-off among all decoder types}. Accordingly, we adopt MLP as the default decoder in \textit{SatelliteCalculator} and further analyze its depth sensitivity in the following section.

\begin{table*}[ht]
    \centering
    \caption{Comparison of decoder types across all tasks. \textbf{\textcolor{best}{Best}} and \textbf{\textcolor{second}{second-best}} values are highlighted.}
    \resizebox{1.0\textwidth}{!}{
    \begin{tabular}{cc|cccc|cccc|cccc|cccc}
    \toprule
    \multirow{2}{*}{\textbf{ID}} & \multirow{2}{*}{\textbf{Task}} &
    \multicolumn{4}{c|}{\textbf{MLP}} &
    \multicolumn{4}{c|}{\textbf{U-Net}} &
    \multicolumn{4}{c|}{\textbf{ResNet}} &
    \multicolumn{4}{c}{\textbf{Transformer}} \\
    \cmidrule(lr){3-6} \cmidrule(lr){7-10} \cmidrule(lr){11-14} \cmidrule(lr){15-18}
    & & \textbf{MAE} & \textbf{PSNR} & $\mathbf{R^2}$ & \textbf{RMSE} & 
          \textbf{MAE} & \textbf{PSNR} & $\mathbf{R^2}$ & \textbf{RMSE} & 
          \textbf{MAE} & \textbf{PSNR} & $\mathbf{R^2}$ & \textbf{RMSE} & 
          \textbf{MAE} & \textbf{PSNR} & $\mathbf{R^2}$ & \textbf{RMSE} \\
    \midrule
    1 & NDVI  & 0.05 & \textbf{\textcolor{second}{29.27}} & \textbf{\textcolor{best}{0.85}} & \textbf{\textcolor{best}{0.22}} & \textbf{\textcolor{second}{0.05}} & 28.79 & \textbf{\textcolor{second}{0.82}} & \textbf{\textcolor{second}{0.23}} & 0.05 & \textbf{\textcolor{best}{29.42}} & 0.83 & 0.23 & \textbf{\textcolor{best}{0.05}} & 29.17 & 0.81 & 0.23 \\
    2 & GNDVI & 0.06 & \textbf{\textcolor{second}{28.99}} & \textbf{\textcolor{best}{0.86}} & \textbf{\textcolor{best}{0.22}} & \textbf{\textcolor{second}{0.06}} & 28.75 & \textbf{\textcolor{second}{0.85}} & \textbf{\textcolor{second}{0.23}} & 0.06 & 28.76 & 0.84 & 0.23 & \textbf{\textcolor{best}{0.06}} & \textbf{\textcolor{best}{28.85}} & 0.83 & 0.24 \\
    3 & SAVI  & 0.07 & \textbf{\textcolor{best}{27.24}} & \textbf{\textcolor{best}{0.77}} & \textbf{\textcolor{best}{0.27}} & 0.07 & 26.54 & 0.74 & 0.29 & \textbf{\textcolor{second}{0.06}} & \textbf{\textcolor{second}{27.09}} & \textbf{\textcolor{second}{0.76}} & \textbf{\textcolor{second}{0.28}} & \textbf{\textcolor{best}{0.06}} & 26.99 & 0.76 & 0.28 \\
    4 & EVI   & \textbf{\textcolor{best}{0.23}} & \textbf{\textcolor{second}{21.95}} & \textbf{\textcolor{best}{0.66}} & \textbf{\textcolor{best}{0.51}} & \textbf{\textcolor{second}{0.23}} & \textbf{\textcolor{best}{22.14}} & \textbf{\textcolor{second}{0.65}} & \textbf{\textcolor{second}{0.51}} & 0.23 & 21.84 & 0.63 & 0.51 & 0.23 & 21.86 & 0.62 & 0.52 \\
    5 & NDWI  & \textbf{\textcolor{second}{0.06}} & 27.79 & \textbf{\textcolor{second}{0.80}} & 0.26 & 0.06 & \textbf{\textcolor{second}{27.96}} & \textbf{\textcolor{best}{0.82}} & \textbf{\textcolor{best}{0.25}} & 0.05 & \textbf{\textcolor{best}{28.14}} & 0.80 & \textbf{\textcolor{second}{0.26}} & \textbf{\textcolor{best}{0.05}} & 28.09 & 0.80 & 0.26 \\
    6 & H     & 2.55 & \textbf{\textcolor{second}{21.05}} & \textbf{\textcolor{best}{0.61}} & \textbf{\textcolor{best}{4.02}} & \textbf{\textcolor{best}{2.55}} & 20.93 & \textbf{\textcolor{second}{0.60}} & \textbf{\textcolor{second}{4.03}} & \textbf{\textcolor{second}{2.55}} & 20.99 & 0.59 & 4.03 & 2.55 & \textbf{\textcolor{best}{21.10}} & 0.59 & 4.03 \\
    7 & AGB   & 21.34 & 21.37 & 0.55 & \textbf{\textcolor{second}{26.04}} & \textbf{\textcolor{second}{21.34}} & \textbf{\textcolor{best}{21.53}} & \textbf{\textcolor{best}{0.56}} & \textbf{\textcolor{best}{26.04}} & \textbf{\textcolor{best}{21.34}} & \textbf{\textcolor{second}{21.66}} & 0.55 & 26.04 & 21.34 & 21.40 & 0.54 & 26.04 \\
    8 & CS    & 16.80 & 21.12 & 0.52 & 21.84 & \textbf{\textcolor{second}{16.80}} & 21.13 & \textbf{\textcolor{best}{0.53}} & \textbf{\textcolor{second}{21.84}} & \textbf{\textcolor{best}{16.80}} & \textbf{\textcolor{best}{21.27}} & 0.51 & 21.85 & 16.78 & \textbf{\textcolor{second}{21.27}} & \textbf{\textcolor{second}{0.53}} & \textbf{\textcolor{best}{21.84}} \\
    \bottomrule
    \end{tabular}
    }
    \label{table_decoder_comparison}
\end{table*}

\begin{table}[ht]
    \centering
    \caption{Comparison of different decoder architectures (MLP with 4 layers) in terms of model size, training time, inference time, and GPU memory usage.}
    \begin{tabular}{ccccc}
    \toprule
    \makecell{\textbf{Decoder}\\ \textbf{Type}} & 
    \makecell{\textbf{Total}\\ \textbf{Params (MB)}} & 
    \makecell{\textbf{Train}\\ \textbf{Time (h)}} & 
    \makecell{\textbf{Inference}\\ \textbf{Time (s/image)}} & 
    \makecell{\textbf{Memory}\\ \textbf{(MB)}} \\
    \midrule
    MLP  & \textbf{\textcolor{second}{478.46}} & \textbf{\textcolor{second}{19.51}} & \textbf{\textcolor{best}{0.48}} & \textbf{\textcolor{best}{1566}} \\
    UNet            & 553.81              & 36.35             & 1.14             & 3601 \\
    ResNet          & 501.50              & \textbf{\textcolor{best}{19.49}}             & \textbf{\textcolor{second}{0.54}} & \textbf{\textcolor{second}{1672}} \\
    Transformer     & \textbf{\textcolor{best}{179.80}}              & 40.87             & 1.19             & 2216 \\
    \bottomrule
    \end{tabular}
    \label{table_efficiency}
\end{table}

\subsubsection{Effect of MLP Depth}

\begin{figure}
    \centering
    \includegraphics[width=1.0\linewidth]{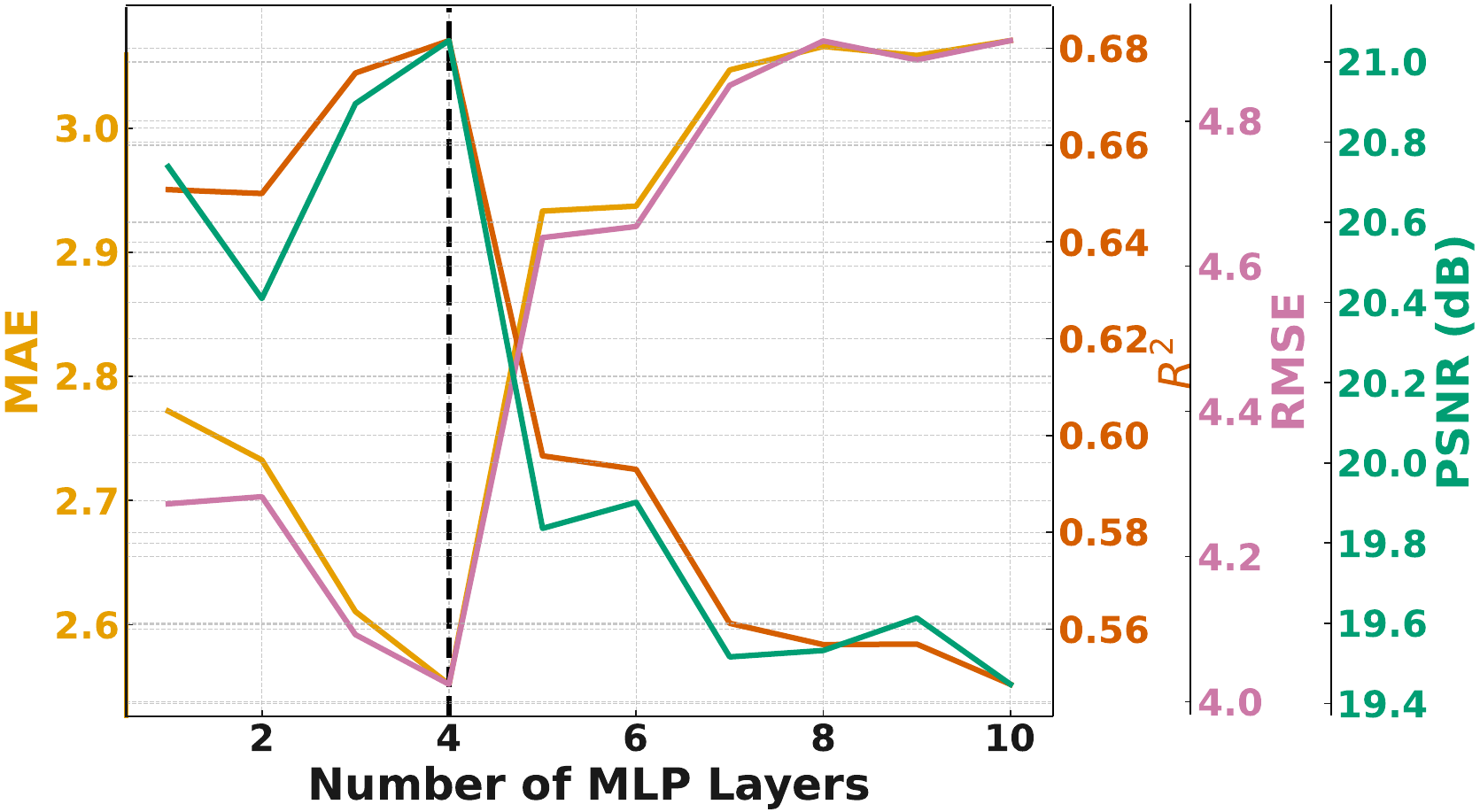}
    \caption{Effect of MLP layer depth on decoder performance. We report the variation of MAE, RMSE, \(R^2\), and PSNR as the MLP depth increases from 1 to 10. The results show that using 4 layers achieves the best overall performance, while deeper architectures tend to overfit and degrade generalization.}
    \label{fig_mlp_layers}
\end{figure}

To evaluate the impact of decoder depth on model performance, we conduct an ablation study by varying the number of layers in the MLP decoder from 1 to 10, while keeping the backbone and all other settings fixed. The results are summarized in Figure~\ref{fig_mlp_layers} and Table~\ref{table_mlp_layers}.

The results show that {using 4 MLP layers yields the best overall performance}, achieving a MAE of {2.55} and an \(R^2\) of {0.68}, with the best balance across all evaluation metrics. Increasing the number of layers beyond 5 leads to noticeable performance degradation: at 10 layers, the RMSE rises to 4.91 and \(R^2\) drops to 0.55, suggesting potential overfitting and reduced generalization capability.

In contrast, shallow configurations (1--2 layers) incur lower computational cost but suffer from higher prediction errors, indicating insufficient representational capacity for modeling the complex non-linear relationships in remote sensing estimation tasks. Considering both accuracy and stability, {we adopt the 4-layer MLP as the default decoder configuration}, achieving a desirable trade-off between efficiency and predictive power.

\section{Conclusion}
\label{sec:conclusion}

We propose \textit{SatelliteCalculator}, the {first vision foundation model} specifically designed for quantitative remote sensing inversion. By leveraging physically defined formulas, we construct a large-scale, label-rich dataset covering eight core ecological variables. Our prompt-guided architecture integrates a frozen Swin Transformer backbone with cross-attentive adapters and lightweight MLP decoders, achieving strong accuracy and efficiency across diverse tasks. The results demonstrate the feasibility of using foundation models for physically interpretable multi-task regression in remote sensing, and provide a scalable, modular framework for future environmental estimation.


\bibliographystyle{IEEEtran}
\bibliography{output}

\newpage
{\appendices

\clearpage
\setcounter{page}{1}
\setcounter{section}{0}
\setcounter{table}{0}
\setcounter{figure}{0}
\setcounter{equation}{0}
\renewcommand{\thetable}{A\arabic{table}}
\renewcommand{\thefigure}{A\arabic{figure}}
\renewcommand{\theequation}{A\arabic{equation}}
\section{Constructing Multi-Task Dataset}
\label{sec_appendix_dataset}




\subsection{Parameter Description}

The spectral indices used in this study are derived from Sentinel-2 Very High-Resolution (VHR) data and serve as key indicators for vegetation condition and ecological structure estimation. All computations are based on four spectral bands: \textbf{Blue} (\textbf{B2}, 490 nm), \textbf{Green} (\textbf{B3}, 560 nm), \textbf{Red} (\textbf{B4}, 665 nm), and \textbf{Near Infrared} (\textbf{NIR}, \textbf{B8}, 842 nm).

We include five vegetation-related indices: NDVI, GNDVI, SAVI, EVI, and NDWI. Among them, SAVI introduces a soil adjustment factor \(L\) to correct background influence, while EVI incorporates gain and aerosol resistance coefficients \(G\), \(C_1\), and \(C_2\) for atmospheric correction. These indices are calculated using standard definitions widely adopted in remote sensing literature.

In addition to indices, we include three structural variables: canopy height (H), aboveground biomass (AGB), and carbon stock (CS), which are estimated from regression models using NDVI, SAVI, and LiDAR-derived height information. Empirical coefficients (\(a, b, c, CF\)) in these models are region-specific and derived from ecological studies on open-canopy forest systems.

\subsection{Spectral Indices}

\begin{enumerate}
    \item \textbf{Normalized Difference Vegetation Index (NDVI)} \cite{Rouse1974}
    \begin{equation}
        NDVI = \frac{B8 - B4}{B8 + B4}
    \end{equation}
    NDVI is widely used to measure vegetation health and biomass productivity. It utilizes the difference between near-infrared (NIR, B8) and red (B4) reflectance to assess chlorophyll activity in plants.

    \item \textbf{Green Normalized Difference Vegetation Index (GNDVI)} \cite{Gitelson1996}
    \begin{equation}
        GNDVI = \frac{B8 - B3}{B8 + B3}
    \end{equation}
    GNDVI is a modification of NDVI that enhances sensitivity to vegetation chlorophyll content by incorporating the green band (B3) instead of the red band (B4).

    \item \textbf{Soil-Adjusted Vegetation Index (SAVI)} \cite{Huete1988}
    \begin{equation}
        SAVI = \frac{(B8 - B4) \times (1 + L)}{B8 + B4 + L}
    \end{equation}
    where \(L = 0.5\) is the soil adjustment factor, which minimizes soil brightness effects in low-vegetation areas.

    \item \textbf{Enhanced Vegetation Index (EVI)} \cite{Huete2002}
    \begin{equation}
        EVI = G \times \frac{(B8 - B4)}{B8 + C_1 \times B4 - C_2 \times B2 + L}
    \end{equation}
    where \( G = 2.5 \), \( C_1 = 6 \), \( C_2 = 7.5 \), and \( L = 1 \).  
    EVI improves upon NDVI by reducing atmospheric and background noise, making it more suitable for dense vegetation monitoring.

    \item \textbf{Normalized Difference Water Index (NDWI)} \cite{McFeeters1996}
    \begin{equation}
        NDWI = \frac{B3 - B8}{B3 + B8}
    \end{equation}
    NDWI is used to monitor water bodies, distinguishing open water from land features. It leverages the high reflectance of water in the green band (B3) and the strong absorption in the NIR band (B8).
\end{enumerate}

\subsection{Ecological Structural Variables}

\begin{enumerate}
    \item \textbf{Canopy Height (H)}
    The canopy height data is directly obtained from the Open-Canopy dataset \cite{fogel2024open} and is not derived from NDVI or any regression model.

    \item \textbf{Aboveground Biomass (AGB)} \cite{Chave2014}
    \begin{equation}
        AGB = a \times H^b
    \end{equation}
    where \( H \) represents canopy height obtained from the Open-Canopy dataset, and \( a, b \) are empirical constants derived from literature or site-specific calibration. 
    
    For temperate forests in France, typical values from literature suggest that for coniferous forests, the empirical coefficients are \( a = 0.118 \) and \( b = 2.53 \); for broadleaf forests, \( a = 0.052 \) and \( b = 2.69 \); and for mixed forests, \( a = 0.067 \) and \( b = 2.58 \) \cite{Paul2019}.

    If forest type is unknown, a general model can be used:
    \begin{equation}
        AGB = 0.067 \times H^{2.58}
    \end{equation}
    which is a widely used allometric equation for estimating biomass in temperate forests.

    \item \textbf{Carbon Stock (CS)} \cite{IPCC2006}
    \begin{equation}
        CS = AGB \times CF
    \end{equation}
    where \( CS \) represents the carbon stock (tC/ha), which is the total amount of carbon stored in aboveground biomass. \( AGB \) refers to the aboveground biomass (t/ha), estimated from canopy height using empirical models. \( CF = 0.47 \) is the carbon fraction factor, indicating that approximately 47\% of the total biomass is composed of carbon, as suggested by IPCC guidelines. This factor may vary depending on tree species and environmental conditions but is commonly used in temperate and tropical forest studies.
\end{enumerate}

\section{Decoder Architecture}

\begin{table}[ht]
\centering
\caption{Architecture of the task-specific MLP head.}
\begin{tabular}{lcc}
\toprule
\textbf{Layer (type)} & \textbf{Output Shape} & \textbf{Activation} \\
\midrule
Linear-1      & [$B$, 1024] & ReLU \\
Linear-2      & [$B$, 1024] & ReLU \\
Linear-3      & [$B$, 1024] & ReLU \\
Linear-4      & [$B$, $H \times W$] & -- \\
\bottomrule
\end{tabular}
\label{tab_mlp_task_head}
\end{table}


\begin{table}[ht]
\centering
\caption{Architecture of the task-specific U-Net regression head.}
\begin{tabular}{lcc}
\toprule
\textbf{Layer (type)} & \textbf{Output Shape} & \textbf{Activation} \\
\midrule
Conv2d-1             & [$B$, 256, $H$, $W$]         & ReLU \\
Conv2d-2             & [$B$, 256, $H$, $W$]         & ReLU \\
MaxPool2d-1          & [$B$, 256, $H/2$, $W/2$]     & -- \\
Conv2d-3             & [$B$, 512, $H/2$, $W/2$]     & ReLU \\
Conv2d-4             & [$B$, 512, $H/2$, $W/2$]     & ReLU \\
MaxPool2d-2          & [$B$, 512, $H/4$, $W/4$]     & -- \\
Conv2d-5             & [$B$, 1024, $H/4$, $W/4$]    & ReLU \\
Conv2d-6             & [$B$, 1024, $H/4$, $W/4$]    & ReLU \\
MaxPool2d-3          & [$B$, 1024, $H/8$, $W/8$]    & -- \\
Conv2d-7             & [$B$, 1024, $H/8$, $W/8$]    & ReLU \\
Conv2d-8             & [$B$, 1024, $H/8$, $W/8$]    & ReLU \\
ConvTranspose2d-1    & [$B$, 512, $H/4$, $W/4$]     & -- \\
Conv2d-9             & [$B$, 512, $H/4$, $W/4$]     & ReLU \\
Conv2d-10            & [$B$, 512, $H/4$, $W/4$]     & ReLU \\
ConvTranspose2d-2    & [$B$, 256, $H/2$, $W/2$]     & -- \\
Conv2d-11            & [$B$, 256, $H/2$, $W/2$]     & ReLU \\
Conv2d-12            & [$B$, 256, $H/2$, $W/2$]     & ReLU \\
ConvTranspose2d-3    & [$B$, 128, $H$, $W$]         & -- \\
Conv2d-13            & [$B$, 128, $H$, $W$]         & ReLU \\
Conv2d-14            & [$B$, 128, $H$, $W$]         & ReLU \\
Conv2d-15            & [$B$, 1, $H$, $W$]           & -- \\
\bottomrule
\end{tabular}
\label{tab_unet_task_head}
\end{table}

\begin{table}[ht]
\centering
\caption{Architecture of the task-specific ResNet regression head.}
\begin{tabular}{lcc}
\toprule
\textbf{Layer (type)} & \textbf{Output Shape} & \textbf{Activation} \\
\midrule
Conv2d-1 (Input Projection)     & [$B$, 1024, $H$, $W$]   & -- \\
ResidualBlock-1                 & [$B$, 1024, $H$, $W$]   & ReLU \\
ResidualBlock-2                 & [$B$, 1024, $H$, $W$]   & ReLU \\
ResidualBlock-3                 & [$B$, 1024, $H$, $W$]   & ReLU \\
Conv2d-2 (Output Projection)    & [$B$, 1, $H$, $W$]      & -- \\
Bilinear Upsampling             & [$B$, 1, 224, 224]      & -- \\
\bottomrule
\end{tabular}
\label{tab_resnet_task_head}
\end{table}

\begin{table}[ht]
\centering
\caption{Architecture of the task-specific Transformer regression head.}
\begin{tabular}{lcc}
\toprule
\textbf{Layer (type)} & \textbf{Output Shape} & \textbf{Activation} \\
\midrule
Position Embedding        & [$B$, $H \times W$, 768]        & -- \\
TransformerEncoder-1      & [$B$, $H \times W$, 768]        & -- \\
TransformerEncoder-2      & [$B$, $H \times W$, 768]        & -- \\
Linear-1 (MLP)            & [$B$, $H \times W$, 1024]       & ReLU \\
Linear-2 (MLP)            & [$B$, $H \times W$, 1]          & -- \\
Reshape + Interpolate     & [$B$, 1, 224, 224]              & -- \\
\bottomrule
\end{tabular}
\label{tab_transformer_task_head}
\end{table}

\section{Figures and Tables}
\label{sec:figures_tables}



\begin{table}[ht]
    \centering
    \caption{Ablation study on the number of MLP layers in the task-specific decoder. We compare performance across MAE, PSNR, \(\mathrm{R^2}\), and RMSE. The 4-layer configuration achieves the best overall results.}
    \begin{tabular}{ccccc}
    \toprule
    \textbf{Layers} & \textbf{MAE}$\downarrow$ & \textbf{PSNR}$\uparrow$ & \(\mathbf{R^2}\)$\uparrow$ & \textbf{RMSE}$\downarrow$ \\
    \midrule
    1  & 2.7720 & 20.7387 & 0.6508 & 4.2726 \\
    2  & 2.7326 & 20.4098 & 0.6500 & 4.2824 \\
    3  & \textbf{\textcolor{second}{2.6105}} & \textbf{\textcolor{second}{20.8959}} & \textbf{\textcolor{second}{0.6749}} & \textbf{\textcolor{second}{4.0923}} \\
    4  & \textbf{\textcolor{best}{2.5519}} & \textbf{\textcolor{best}{21.0532}} & \textbf{\textcolor{best}{0.6816}} & \textbf{\textcolor{best}{4.0239}} \\
    5  & 2.9332 & 19.8367 & 0.5958 & 4.6394 \\
    6  & 2.9372 & 19.9017 & 0.5930 & 4.6547 \\
    7  & 3.0469 & 19.5162 & 0.5612 & 4.8492 \\
    8  & 3.0659 & 19.5322 & 0.5568 & 4.9104 \\
    9  & 3.0586 & 19.6133 & 0.5569 & 4.8843 \\
    10 & 3.0706 & 19.4479 & 0.5486 & 4.9109 \\
    \bottomrule
    \end{tabular}
    \label{table_mlp_layers}
\end{table}


}

\newpage

 




\vfill

\end{document}